\documentclass[preprints,article,accept,pdftex,moreauthors]{Definitions/mdpi} 

\firstpage{1} 
\pubvolume{1}
\issuenum{1}
\articlenumber{0}
\pubyear{2026}
\copyrightyear{2026}
\externaleditor{Kim Andrea Nicoli, Lei Wang and Anindita Maiti} 
\datereceived{28 December 2025} 
\daterevised{30 January 2026} 
\dateaccepted{3 February 2026} 
\datepublished{ } 
\pdfoutput=1 

\Title{The 
 Information Dynamics of Generative Diffusion}

\def\rvx{{\mathbf{x}}}
\def\rvy{{\mathbf{y}}}
\def\rvz{{\mathbf{z}}}

\DeclareMathOperator{\Var}{Var}
\DeclareMathOperator*{\argmin}{arg\,min}

\usepackage{amsmath}
\usepackage{graphicx}    
\usepackage{subcaption}  

\newcommand{\vect}[1]{\boldsymbol{#1}}
\newcommand{\mean}[2]{\mathbb{E}_{#2}\left[ #1 \right]}

\newcommand{\x}[1]{\vect{x}_{#1}} 
\newcommand{\y}{\vect{y}}
\newcommand{\score}[1]{\nabla \log{p_{#1}(\x{#1})}}

\newcommand{\norm}[1]{\left\lVert#1\right\rVert}

\Author{Dejan Stančević 
 * and Luca Ambrogioni *}

\AuthorNames{Dejan Stančević, Luca Ambrogioni}

\address[1]{%
Donders Institute for Brain, Cognition and Behaviour, 
Radboud University, 6525 GD Nijmegen, The Netherlands 
}

\corres{\hangafter=1 \hangindent=1.05em \hspace{-0.82em}Correspondence: dejan.stancevic@donders.ru.nl (D.S.); luca.ambrogioni@donders.ru.nl (L.A.)}

\abstract{Generative diffusion models have emerged as a powerful class of models in machine learning, yet a unified theoretical understanding of their operation is still developing. This paper provides an integrated perspective on generative diffusion by connecting the information-theoretic, dynamical, and thermodynamic aspects. We demonstrate that the rate of conditional entropy production during generation (i.e., the generative bandwidth) is directly governed by the expected divergence of the score function's vector field. This divergence, in turn, is linked to the branching of trajectories and generative bifurcations, which we characterize as symmetry-breaking phase transitions in the energy landscape. Beyond ensemble averages, we demonstrate that symmetry-breaking decisions are revealed by peaks in the variance of pathwise conditional entropy, capturing heterogeneity in how individual trajectories resolve uncertainty. Together, these results establish generative diffusion as a process of controlled, noise-induced symmetry breaking, in which the score function acts as a dynamic nonlinear filter that regulates both the rate and variability of information flow from noise to data.}


\makeatletter
\fancypagestyle{plain}{\fancyhf{}\fancyfoot[C]{\thepage}}
\makeatother
\begin{document}

\section{Introduction}
Generative diffusion models have rapidly become one of the most successful frameworks for high-dimensional generation across images, audio, and~video \citep{sohldickstein2015deep,ho2020denoising,dhariwal2021diffusion,kong2021diffwave,singer2022makeavideo}. In~practice, they synthesize samples by iteratively denoising from simple noise, and~they support flexible conditioning and editing via guidance mechanisms that trade off fidelity and diversity~\citep{dhariwal2021diffusion,ho2022classifierfree,rombach2022latent}. They were introduced in \citet{sohldickstein2015deep} in analogy with stochastic thermodynamics and~later unified under score-based/SDE formulations that clarify training objectives and reverse-time sampling \citep{song2021score,ho2020denoising,song2022denoising} (the diffusion formalism, notation, and~the SDE viewpoint can be found in Section~\ref{section: diffusion models}). Despite these efforts, a~unified conceptual understanding of their behavior is still emerging. Several perspectives on information theory, stochastic thermodynamics, and~the statistical physics of symmetry breaking have each shed light on different aspects of diffusion models, but~their interrelations remain fragmented. The~purpose of this perspective paper is to integrate these viewpoints into a single coherent theoretical~picture.

Our central thesis is that generation 
in diffusion models proceeds through a sequence of
noise-driven symmetry-breaking transitions. These transitions determine when and how the model
commits to a specific generative outcome, structure the flow of information, regulate entropy
production, and~shape the geometry of trajectories in state space. We refer to this synthesis as the
information thermodynamics of generative diffusion.

\begin{enumerate}[label=,leftmargin=0em,labelsep=0mm]
\item{Information 
theory and entropy-based perspectives}
\end{enumerate}

A growing line of work has examined diffusion models from the standpoint of information theory,
focusing especially on how information about the clean sample $x_0$ is progressively revealed as
noise is removed. Recent works have proposed information-theoretic decompositions of diffusion
dynamics~\citep{kong2023information,kong2023interpretable} and have explored the role of conditional
entropy in designing improved training and sampling schedules
\citep{stancevic2025entropic,dieleman2022continuous}. Furthermore, \citet{franzese2025latent} show how information-theoretic tools reveal the mechanisms by which latent abstractions guide generation. These approaches treat diffusion as a
sequential information transfer process and highlight that the effectiveness of generation depends
on how rapidly uncertainty about $x_0$ can be reduced. Central to these results is the observation
that the conditional entropy rate is directly linked to geometric quantities such as the divergence
of the score and the curvature of the log-density. This suggests that information flow is deeply
connected to the underlying dynamical and geometric structure of the generative~process.

\begin{enumerate}[label=,leftmargin=0em,labelsep=0mm]
\item{Phase transitions, associative memories, and~symmetry breaking}
\end{enumerate}

\textls[-35]{Parallel developments in statistical physics have revealed that diffusion models exhibit noise-driven symmetry-breaking events, where the score field undergoes bifurcations and the generative trajectories split into distinct modes \citep{raya2023spontaneous,ambrogioni2024statistical}. High-dimensional analyses have linked these transitions to mean-field phase transitions~\citep{biroli2023generative} and to dynamical behavior captured by stochastic localization~\citep{alaoui2023sampling,huang2024sampling,montanari2023sampling,benton2024nearly}. These bifurcations correlate with sharp changes in the Hessian of the log-density, revealing a connection between symmetry breaking and information geometry. Similar mechanisms have been studied in hierarchical generative settings~\citep{sclocchi2025phase,sclocchi2025probing} and in analyses of memorization, mode formation, and~semantic emergence~\citep{biroli2024dynamical,bonnaire2025diffusion,achilli2025memorization,achilli2024losing}. Generative diffusion models have also been directly connected to modern Hopfield networks and other associative memory networks~\cite{ambrogioni2024search, hoover2023memory, hess2025associative, jeon2024understanding}, whereas generalization has been associated with the emergence of spurious states \citep{pham2025memorization}. Across these domains, the~key unifying insight is that the Hessian (or  Jacobian) score mediates both stability of generative trajectories and the structure of the data~manifold.}

\begin{enumerate}[label=,leftmargin=0em,labelsep=0mm]
\item{Thermodynamics and the role of entropy}
\end{enumerate}

The connection between diffusion models and stochastic thermodynamics was first made explicit in \citet{sohldickstein2015deep}, motivating a thermodynamic view of generation. Furthermore, this connection was strengthened with a mathematical framework based on stochastic differential equations (SDEs) formulated \mbox{in~\citet{song2021score, rombach2022latent}} and is central to the modern understanding of diffusion models. However, the~notion of entropy that is commonly used in stochastic thermodynamics~\citep{premkumar2024neural, seifert2005entropy} measures the irreversibility of the forward process. While such quantities yield elegant speed--accuracy tradeoffs \citep{ikeda2025speed}, they characterize the evolution of the distribution of trajectories rather than the uncertainty relevant for generating a single sample. Instead, we argue that what matters during generation is the uncertainty about the clean sample. 

\begin{enumerate}[label=,leftmargin=0em,labelsep=0mm]
\item{Contributions}
\end{enumerate}

We synthesize several existing lines of work on information flow, symmetry breaking, and~instability in diffusion models into a single coherent narrative, and~we clarify how these viewpoints fit together. The~main new technical contributions appear in \mbox{Sections~\ref{Jacobian dynamics} and \ref{Variance peak}}, where we connect entropy rate to trajectory divergence and analyze a variance-based signature of decision points (including its speciation-time scaling), respectively. Concretely, we provide the following:

\begin{enumerate}[labelsep=5mm]
    \item Entropy production as a signature of symmetry breaking: 
We review the known expression for conditional entropy production in diffusion dynamics and provide an intuitive interpretation (see \citep{stancevic2025entropic, kong2023information} for a practical use of the entropy production expression). We then connect this information-theoretic quantity to symmetry-breaking phenomena reported in \citep{raya2023spontaneous,ambrogioni2024statistical}, emphasizing that bifurcations manifest as pronounced, ensemble-level changes in information measures. In~particular, symmetry breaking induces a transient loss of identifiability of $x_0$, which appears as a peak in the conditional entropy~rate.
    
\item \textls[-15]{Noise-driven decisions via posterior geometry: We interpret these entropy-rate signatures through the geometry of the score field. Near~low-curvature directions, the~score weakens and temporarily loses its ability to suppress stochastic fluctuations, so noise becomes the effective selector of the branch that the generative trajectory follows. This provides a unified, mechanism-level explanation for ``decision'' events during~generation.}
    
\item Local divergence of trajectories under reverse-time dynamics: We show that the same loss of curvature is reflected in the local linearization of the reverse-time flow. The~Jacobian of the score develops expanding directions, implying local exponential amplification of small differences between nearby generative trajectories over finite horizons. This explains how noise fluctuations can propagate and shape the final generative~outcome.
    
\item Variance peak as a speciation-time marker: Motivated by tools from stochastic thermodynamics, we introduce the variance of pathwise conditional entropy. Along individual trajectories, the~pathwise conditional entropy need not decrease monotonically and may transiently increase, reflecting temporary ambiguity in the resolution of uncertainty about $x_0$. This trajectory heterogeneity produces a pronounced peak in the variance, and~we show how this peak concentrates around the speciation time \citep{biroli2024dynamical}.
\end{enumerate}

Together, these results organize existing insights into a unified picture in which entropy production and local stability provide complementary perspectives on the same noise-driven branching mechanism. At~the same time, our variance-based analysis offers a new lens on trajectory-level generative dynamics, highlighting a form of pathwise heterogeneity and decision-making that is fundamentally absent in autoregressive~generation.


\section{Information~Theory}
We start by presenting an introduction to the information theory of sequential generative modeling, which will open the door to the analysis of generative~diffusion. 

Consider a game of Twenty Questions where an interrogator player may ask twenty binary questions concerning a set to an "oracle" player in order to gradually reveal the identity of a predetermined element $\rvy^*$ in a finite set $\Omega = \{\rvy_1, \dots, \rvy_{N_0}\}$ with $N_0$ elements. We denote the size of the possible set $\Omega_{j-1}(a_{1:j-1})$ after $j-1$ questions as $N_{j-1}(a_{1:j-1})$. The~answer $a_j$ to the $j$-th question $q_j$ then divides the set into two possible subsets with sizes $N^1_j(a_{1:j-1}) = N_j(a_j = 1, a_{1:j-1})$ and $N^0_j(a_{1:j-1}) = N_{j-1}(a_{1:j-1}) - N_j(a_j = 1, a_{1:j-1})$. Assuming a fixed set of questions, the~expected uncertainty experienced by the player after the $j$-th question can be quantified by the conditional entropy: 
\begin{equation}
    \mathbf{H}[\rvy^* \mid a_{1:j}] = - \mean{ \log_2{p(\rvy \mid a_{1:j})}}{\rvy^*, a_{1:j}} = \mean{\log_2{N_{j}(a_{1:j})}}{a_{1:j}}
\end{equation}
where $\rvy^*$ is sampled uniformly from $\Omega$. Under~these conditions, the~expected entropy reduction associated to a given question is given by
\begin{equation} \label{eq: discrete entropy}
    \Delta \mathbf{H}_j = \mean{\log_2{N_{j-1}} - \frac{N_j^0}{N_{j-1}} \log_2{N_j^0} + \frac{N_j^1}{N_{j-1}} \log_2{N_j^1}}{a_{1:j}}~,
\end{equation}
where we left the dependence on the set of answers implicit to unclutter the notation. It is easy to see that the maximum bit rate is $1$, which is achieved when $N^0_j = N_{j-1}/2$. Assuming that $20$ questions are enough to fully identify the value of $\rvy^*$, we can encode each $\rvy$ in the string of binary values $a_{1:20}$, which makes clear that the question-answering process consists of gradually filling in this string. Using the language of generative diffusion, we can re-frame this process in terms of a 'forward' process, where the string $a_{1:20}$ corresponding to an element of $\Omega$ is sampled in advance and then transmitted to the $j$-th 'time point' through the following non-injective forward process
\begin{equation}
    R_j(a_{1:20}) = a_{1:j}~,
\end{equation}
which deterministically suppresses information by masking the values of the string. The~solution to a Twenty Question game can then be seen as inverting this 'forward process'. Note that the forward process leads to a sequence of monotonically non-decreasing marginal conditional entropies $\mathbf{H}(\rvy^* \mid a_{1:j}) < \mathbf{H}(\rvy^* \mid a_{1:{j-1}})$, which is a fundamental feature of a forward process in diffusion models that captures the fact that information is lost by the forward~transformation. 

Now consider the case where a lazy oracle forgot to select a word in advance and decides instead to answer the questions at random under the probability determined by the sizes $N^0_j$ and $N^1_j$, which we assume to be fixed given the questions. Strikingly, this reformulation does not make any observable difference from the point of view of the interrogator as each (randomly sampled) answer equally reduces the space of possible words and it results in the same entropy reduction, until~a final guess can be offered. Therefore, the~game of Twenty Questions with a random oracle can be interpreted as a sequential generative process where the state at 'time' $j$ is given by a binary string $a_{1:j}$ with Markov transition probabilities
\begin{equation}
    p(a_{j+1} = 0 \mid a_{1:j}) = \frac{N^0_{j+1}(a_{1:j})}{N_{j}(a_{1:j})}
\end{equation}

The conditional entropy rate $\Delta H_j$ determines how much information is transferred from 'time' $j$ to the final~generation.

\textls[-15]{As we shall see, the~reverse diffusion process can be seen as analogous to this 'generative game' with the score function playing the part of the interrogator and the noise $\boldsymbol{\epsilon}_t$ playing the role of the oracle. Like in the interrogator in the generative Twenty Questions game, the~score function can reduce the information transfer by tilting the probabilities of the stochastic increments out of uniformity, which reduces the impact of the noise. This phenomenon is related to the divergence of the vector field induced by the score function, which amplifies small perturbations during generative dynamics. We will also see that the phenomenon is connected to the branching of paths of fixed points of the score and consequently to the phenomenon of generative phase transitions and spontaneous symmetry breaking \citep{raya2023spontaneous}. }

\section{Diffusion~Models}
\label{section: diffusion models}

The sequential generation example outlined above is analogous to the masked diffusion models~\cite{lou2023discrete, sahoo2024simple}. On~the other hand, continuous diffusion models cast generation as the time reversal of a noising process that transports a complex data distribution $p_0$ to a simple reference distribution (typically an isotropic Gaussian) through a sequence of progressively corrupted marginals $\{p_t\}_{t\in[0,T]}$ \citep{sohldickstein2015deep, ho2020denoising,song2021score}. In~continuous time, the~forward process is specified by an SDE that gradually destroys information about $X_0$ by injecting noise, while the reverse-time process progressively restores structure. Therefore, intermediate states $X_t$ can be interpreted as partially noised versions of the~data.

Formally, an~SDE specifies the evolution of a random variable as
\begin{equation}
    dX_t = f(X_t,t)\,dt + g(t)\,dW_t,
\end{equation}
where $W_t$ is a standard Wiener process, $f$ is a drift field, and~$g$ controls the noise scale. The~associated marginal densities $p_t(x)$ evolve according to the Fokker--Planck equation
\begin{equation}
\label{eq: fokker_planck}
    \partial_t p_t(x) 
    = \sum_{j=1}^d \partial_{x_j} 
    \left( - f_j(x, t)
    + \frac{g^2(t)}{2} \partial_{x_j} \right) 
    p_t(x).
\end{equation}

Different diffusion formulations correspond to different choices of $(f,g)$, or~equivalently, different parameterizations of Gaussian perturbation kernels $p(x_t\mid x_0)$. For~the standard choices used in diffusion models, the~forward noising admits a closed-form Gaussian kernel, which we write as
\begin{equation}
\label{eq: forward_kernel}
    p(x_t\mid x_0) \;=\; \mathcal{N}\!\big(x_t;\,\alpha_t x_0,\,\sigma_t^2 I\big),
\end{equation}
with $\alpha_t$ controlling the attenuation of the signal and $\sigma_t$ the noise~level.

\textls[25]{In the variance-preserving (VP) setting, the~forward process is the Ornstein--Uhlenbeck process,}
\begin{equation}
    dX_t = -\tfrac12\beta(t)X_t\,dt + \sqrt{\beta(t)}\,dW_t,
\end{equation}
which \textls[25]{implies the closed-form corruption $X_t=\alpha_t X_0+\sigma_t Z$ with $Z\sim\mathcal{N}(0,I)$ and $\alpha_t^2+\sigma_t^2=1$ \citep{ho2020denoising,song2021score}.}

\textls[-25]{In the variance-exploding (VE) setting, the~drift is set to zero and the noise scale~increases,}
\begin{equation}
    dX_t = g(t)\,dW_t
    \qquad\Longleftrightarrow\qquad
    X_t = X_0 + \sigma_t Z,
\end{equation}
so $\mathrm{Var}(X_t)$ grows with $t$ \citep{song2021score}.

Finally, the~EDM parameterization \citep{karras2022elucidating} can be viewed as a VE-type SDE written directly in terms of the noise level $\sigma$ (rather than physical time), with~the canonical choice $\sigma_t=t$ and $X_t = X_0 + t Z$. In~this paper, we will mostly adopt this EDM convention for notational convenience. The~main identities translate broadly across VP/VE/EDM formulations. When the specific choice of forward process becomes essential, notably in our discussion of sharp speciation-time behavior and the variance peak, we will make this explicit (there, we focus on the VP/OU setting).

Generation is obtained by reversing the forward process. Reversing an SDE introduces an additional drift term involving the score $\nabla_x\log p_t(x)$ \citep{anderson1982reverse},
\begin{equation}
\label{eq: backward_sde}
    dX_t = \left(f(X_t,t) - g^2(t)\nabla_x\log p_t(X_t)\right)dt + g(t)\,d\widetilde{W}_t.
\end{equation}
{The fundamental} mathematical object that determines the reverse dynamics is the score function, which in this case can be expressed as
$
    \score{t} = \mean{\frac{\y - \x{t}}{\sigma^2(t)}}{\y \mid \x{t}},
$
where the expectation is taken with respect to the conditional distribution $p(\rvy \mid \x{t}) \propto p(\x{t} \mid \rvy) \rho(\rvy)$. This expression can be further simplified by noticing that $\x{t} = \rvy + \sigma(t) \rvz_t$:
\begin{equation} \label{eq: score}
    \score{t} = - \mean{\frac{\rvz_t}{\sigma(t)}}{\rvz_t \mid \x{t}}
\end{equation}
where $\rvz$ is a standard normal vector. In~other words, the~score is the negative of the average (rescaled) noise, and it therefore provides the optimal (infinitesimal) denoising~direction.

If the score were known exactly, sampling \eqref{eq: backward_sde} from the (approximately Gaussian) terminal distribution of the forward process would produce exact samples from $p_0$. However, in~practice, the~score is not available and is approximated by a neural network $s(\rvx_{t}; \theta)$ trained by (denoising) score matching \citep{hyvärinen2005estimation, vincent2011score, song2021score}. In~other words, it is trained to minimize
\begin{equation} \label{eq: score matching}
    \mathcal{L}_{\text{sm}}(\theta, t) = \mean{\norm{\sigma(t) \score{t} - s(\x{t}; \theta) }^2}{\x{t}} 
\end{equation}
{This} loss function cannot be computed directly because the true score is not available. However, Equation~(\ref{eq: score matching}) can be re-written using Equation~(\ref{eq: score}) and expanding the square:
{\small\begin{align} \label{eq: denoising score matching}
    \mathcal{L}_{\text{sm}}(\theta, t) &= \mean{\norm{\mean{\rvz_t}{\rvz_t \mid \x{t}}~ + s(\x{t}; \theta)}^2}{\x{t}} \\ \nonumber
    &= \mean{\norm{\rvz_t + s(\y + \sigma(t) \rvz_t; \theta) }^2}{\rvz_t, \rvy} - \mean{\norm{{\rvz_t} + \sigma(t) \nabla \log{p_t(\y + \sigma(t) \rvz_t)} }^2}{\rvz_t, \rvy}~.
\end{align}}
{Note} that the second term is constant in $\theta$, which means that the gradient solely depends on the denoising loss:
\begin{align}
    \mathcal{L}_{\text{d}}(\theta, t) &= \mean{\norm{\rvz_t + s(\y + \sigma(t) \rvz_t; \theta) }^2}{\rvz_t, \rvy}~.
\end{align}
{The} constant term 
\begin{equation}
    C_t = \mean{\norm{{\rvz_t} + \sigma(t) \nabla \log{p_t(\y + \sigma(t) \rvz_t)} }^2}{\rvz_t, \rvy}
\end{equation}
is of high importance for our current purposes. It quantifies the loss of the denoiser obtained from the score function. This is therefore the unavoidable part of the denoising error that is still present given a perfectly trained network. With~a few manipulations, it is possible to show that this term is in fact equal to the variance of the posterior denoising distribution:
\begin{equation}
    C_t = \mean{\text{var}(\y \mid \x{t})}{\y, \x{t}}~,
\end{equation}
which allows us to interpret this term as a measure of uncertainty at time $t$ on the final outcome of the generative~trajectory.

In contrast, throughout this paper, we analyze the oracle (perfect-score) dynamics, assuming access to the exact score $\nabla_x\log p_t(x)$ in order to isolate intrinsic information-theoretic and dynamical mechanisms of the generative process. In~practical models, approximation error means that equalities derived under the oracle assumption may only hold approximately. Moreover, the~learned vector field used in sampling need not coincide with the gradient of any true log-density (i.e., it may be non-integrable), which can introduce inconsistencies relative to the idealized score-driven~dynamics.

\section{Generative Information Transfer in Score Matching~Diffusion}

To characterize the generative information transfer we need to compute the conditional entropy rate $\dot{\mathbf{H}}[\y \mid \x{t}]$, which is   analogous to the discrete entropy reduction we gave in Equation~(\ref{eq: discrete entropy}). The~conditional entropy is defined as
\begin{equation} \label{eq: conditional entropy}
    \mathbf{H}[\y \mid \x{t}] = -\mean{\log{p(\y \mid \x{t})}}{\y, \x{t}}
\end{equation}
{To} find the entropy rate, we can take the temporal derivative of Equation~(\ref{eq: conditional entropy}) and use the Fokker--Planck equation, which in our case is just the heat equation:
\begin{equation}
    \partial_t p_t(\x{t}) = \frac{1}{2}\nu^2(t) \nabla^2 p_t(\x{t})~.
\end{equation}
{Using} integration by parts, this results in
\begin{equation}
\begin{aligned}
    \dot{\mathbf{H}}[\y|\x{t}] &= \frac{\nu^2(t)}{2} \left( \mathbb{E}_{p(\x{t}, \x{0})} [\norm{\nabla \log p(\x{t}|\x{0})}^2 ] - \mathbb{E}_{p_t(\x{t})} [\norm{ \nabla \log p(\x{t})}^2 ] \right) \\
    &= \frac{\nu^2(t)}{2} \left( \frac{D}{\sigma^2(t)} - \mathbb{E}_{p_t(\x{t})} [\norm{ \nabla \log p(\x{t})}^2 ] \right),
\end{aligned}
\label{eq:entropy rate formula}
\end{equation}
where $D$ is the dimensionality of the ambient space. From~this formula, we can see that the maximal bandwidth is reached when the Euclidean norm of the score function is~minimized.

\subsection{Score Norm and Posterior~Concentration}

To gain some insight into the significance of the square norm and the expression for the conditional entropy, we will consider the following case. We assume a discrete data distribution
\(
p_0(y)=\frac{1}{N}\sum_{i=1}^N \delta(y-y_i)
\)
with an empirical mean equal to~zero.

At time $t$, the~marginal density is given by a Gaussian smoothing of the data,
\begin{equation}
p_t(x)=\frac{1}{N}\sum_{i=1}^N \varphi_{\sigma(t)}(x-y_i),
\end{equation}
where $\varphi_{\sigma(t)}$ denotes an isotropic Gaussian with variance $\sigma^2(t)$. The~posterior distribution over datapoints is
\begin{equation}
p(y_i\mid x_t)=\frac{\varphi_{\sigma(t)}(x_t-y_i)}{\sum_{k=1}^N \varphi_{\sigma(t)}(x_t-y_k)} .
\end{equation}
{The} score function can then be written as the posterior average
\begin{equation}
\nabla\log p_t(x_t)
=
\mathbb E_{y\mid x_t}\!\left[\frac{y-x_t}{\sigma^2(t)}\right]
=
\frac{1}{\sigma^2(t)}\Big(\mu(x_t)-x_t\Big),
\qquad
\mu(x_t):=\mathbb E[y\mid x_t].
\label{eq:score_posterior_mean}
\end{equation}

In addition, we assume that the data vectors satisfy
\begin{equation}
y_i^\top y_j \approx 0 \quad (i\neq j),
\qquad
\|y_i\|^2 \approx R^2 ,
\label{eq:orthogonal_data}
\end{equation}
i.e., datapoints are approximately orthogonal and lie at a common distance $R$ from the mean. Even though this holds only in special cases (e.g., randomly sampled data points from a Gaussian in high dimension) and~is not representative of the commonly encountered datasets, we proceed with the setup as an intuitive approach to understanding the meaning of the norm of the score and its role in the entropy expression (Equation (\ref{eq:entropy rate formula})).

Under these assumptions, the~squared norm of the posterior mean simplifies to
\begin{equation}
\|\mu(x_t)\|^2
=
\Big\|\sum_{i=1}^N p(y_i \mid x_t)\,y_i\Big\|^2
\approx
R^2 \sum_{i=1}^N p(y_i \mid x_t)^2 .
\label{eq:mu_norm_purity}
\end{equation}

Taking expectations with respect to \(p_t(x_t)\), we obtain
\begin{equation}
\mathbb E_{x_t}\!\left[\|\nabla\log p_t(x_t)\|^2\right] = \frac{1}{\sigma^4(t)} \Big( \mathbb E_{x_t}\!\left[\|\mu(x_t)\|^2\right]
- 2\,\mathbb E_{x_t}[x_t^\top \mu(x_t)] + \mathbb E_{x_t}[\|x_t\|^2] \Big).
\label{eq:score_norm_expand}
\end{equation}
The first term captures the data-dependent structure of the score and, using
\mbox{Equation~\eqref{eq:mu_norm_purity}}, can be written as
\begin{equation}
\mathbb E_{x_t}\!\left[\|\mu(x_t)\|^2\right]
\approx
R^2\,
\mathbb E_{x_t}\!\left[\sum_{i=1}^N p(y_i \mid x_t)^2\right].
\label{eq:expected_purity}
\end{equation}
{The} quantity
\(
\sum_i p(y_i \mid x_t)^2
\)
measures the concentration of the posterior over datapoints. It satisfies \(1/N\leq \sum_i p(y_i \mid x_t)^2\leq 1\), interpolating between a fully diffuse posterior and complete concentration on a single~datapoint.

The remaining two terms in Equation~\eqref{eq:score_norm_expand} can be estimated explicitly under the forward model \(x_t = y + \sigma(t) z\), where \(z\sim\mathcal N(0,I)\) is
independent of \(y\). We have
\begin{equation}
\mathbb E_{x_t}[x_t^\top \mu(x_t)]
=
\mathbb E_{x_t}\!\big[x_t^\top \mathbb E[y\mid x_t]\big]
=
\mathbb E_{x_t,y}[x_t^\top y]
=
\mathbb E\|y\|^2
\approx R^2 ,
\label{eq:cross_term}
\end{equation}
\begin{equation}
\mathbb E_{x_t}\!\left[\|x_t\|^2\right]
=
\mathbb E\|y\|^2 + \sigma^2(t)\,\mathbb E\|z\|^2
\approx
R^2 + D\sigma^2(t),
\label{eq:x_norm}
\end{equation}
where \(D\) denotes the ambient~dimensionality.

Substituting Equations
~\eqref{eq:expected_purity}--\eqref{eq:x_norm}
into Equation~\eqref{eq:score_norm_expand} yields
\begin{equation}
\mathbb E_{x_t}\!\left[\|\nabla\log p_t(x_t)\|^2\right]
\approx
\frac{R^2}{\sigma^4(t)}
\Bigg(
\mathbb E_{x_t}\!\left[\sum_{i=1}^N p(y_i\mid x_t)^2\right] - 1
\Bigg)
+
\frac{D}{\sigma^2(t)} .
\label{eq:score_norm_final}
\end{equation}

The second term coincides with the expected squared norm of the score of the forward Gaussian kernel and therefore represents a data-independent baseline contribution. The~first term encodes the deviation from pure diffusion induced by the structure of the dataset and depends solely on the posterior distribution over~datapoints.

Using the bound
\(
1/N \leq \sum_{i=1}^N p(y_i\mid x_t)^2 \leq 1
\),
we obtain the inequality
\begin{equation}
- \frac{(N-1)R^2}{N\sigma^4(t)}
\;\leq\;
\frac{R^2}{\sigma^4(t)}
\Bigg(
\mathbb E_{x_t}\!\left[\sum_{i=1}^N p(y_i\mid x_t)^2\right] - 1
\Bigg)
\;\leq\; 0 .
\end{equation}
{As} a consequence, the~expected squared norm of the score is always bounded above by the forward kernel contribution, ensuring that the marginal entropy remains a monotonically increasing function of~time.

Further insight can be gained by rewriting
\begin{equation}
\mathbb E_{x_t}\!\left[\sum_{i=1}^N p(y_i\mid x_t)^2\right]
=
\frac{1}{N}
\sum_{i=1}^N
\int p(y_i\mid x_t)\,p(x_t\mid y_i)\,dx_t .
\label{eq:purity_overlap}
\end{equation}
{This} expression makes explicit that the deviation from the diffusion baseline is controlled by the overlap of the forward kernels. If, at~time $t$, the~datapoints have effectively merged into \(m\) indistinguishable groups (with identical posteriors), the~term evaluates to \(m/N\),~yielding
\begin{equation}
\mathbb E_{x_t}\!\left[\sum_{i=1}^N p(y_i\mid x_t)^2\right] - 1
=
\frac{m-N}{N}.
\label{eq: Mixture approximation}
\end{equation}
{Therefore,} increasing mixing among datapoints (smaller \(m\)) makes the data-dependent term more negative, reducing the expected score norm and increasing the conditional entropy~rate.

This result allows us to interpret the magnitude of the score vector as a quantitative estimate of uncertainty in the denoising process: when multiple datapoints are compatible with the noisy state \(x_t\), posterior averaging suppresses the score, leading to enhanced entropy production (Equation (\ref{eq:entropy rate formula})). As~we shall see in the rest of the paper, we can associate peaks in the entropy rates with symmetry-breaking bifurcations that correspond to noise-induced 'choices' between possible data~points.

\subsection{Conditional Entropy Production as Optimal~Error}
The conditional entropy rate quantifies the instantaneous generative information transfer at any given moment in time. It can be shown (see \citep{stancevic2025entropic}) that this quantity is closely connected to the optimal denoising squared error, which is the variance of the denoising~distribution:
\begin{equation}
    \dot{\mathbf{H}}[\y|\x{t}] = \frac{1}{2} \frac{\nu^2(t)}{ \sigma^4(t)}\mean{\text{var}(\y \mid \x{t})}{\y, \x{t}}~.
\end{equation}
{Intuitively}, this means that the information rate is directly related to the denoising uncertainty at a given~time. 

Using this relation, we can now re-express the denoising score matching formula in Equation~(\ref{eq: denoising score matching}) in terms of the conditional entropy rate:
\begin{equation}
    \mean{\norm{\mean{\rvz}{\rvz_t\mid \x{t}} - s(\x{t}; \theta)}^2}{\x{t}} + \frac{2 \sigma^4(t)}{\nu^2(t)}\dot{\mathbf{H}}[\y|\x{t}]
    = \mean{\norm{\rvz_t- s(\y + \sigma(t) \rvz_t; \theta) }^2}{\rvz_t, \rvy}~,
\end{equation}
which implies that the entropy rate can be estimated from the training loss if we assume that the network is~well-trained. 

\subsection{Generative~Bandwidth}
It is insightful to investigate under what circumstances the score-matching diffusion model can achieve the maximum possible generative bandwidth. From~Equation~(\ref{eq:entropy rate formula}), it is clear that this happens when $\mean{\norm{\score{t}}}{} = 0$, which in turn is obtained if the score vanishes almost~everywhere.  

To realize this situation, we can consider a data distribution $p_h(\y)$ to be a centered multivariate normal with variance $h^2$. In~this case, the~score function is just
\begin{equation}
    \score{t} = -\frac{\x{t}}{\sigma^2(t) + h^2}~,
\end{equation}
which vanishes everywhere for $h \rightarrow \infty$, giving a maximum entropy rate:
\begin{equation}
    \dot{\mathbf{H}}[\x{0} \mid \x{t}] = \frac{1}{2} \frac{D \nu^2(t)}{\sigma^2(t)}~.
\end{equation}
{This} corresponds to a setting where the particles are free to diffuse since every possible generation is equally likely. From~this, we can conclude that the score function has the negative role of suppressing fluctuations along 'unwanted directions' to preserve the statistics of the data and that peaks in the information transfer comes from periods where noise fluctuations are not suppressed. Note that the maximum bandwidth scales with the dimensionality $D$.

Now consider the case where the distribution of the data is a centered Gaussian in a $D_{\text{data}}$-dimensional subspace with $D_{\text{data}} \leq D$. In~this case, the~expected norm of the score decomposes as follows
\begin{equation}
    \mean{\norm{\score{t}}^2}{} = \frac{D_{\text{data}}}{\sigma^2(t) + h^2} + \frac{D - D_{\text{data}}}{\sigma^2(t)}~ \rightarrow \frac{D - D_{\text{data}}}{\sigma^2(t)}~
\end{equation}
which leads to the entropy rate
\begin{equation}\label{eq: manifold entropy rate}
    \dot{\mathbf{H}}[\x{0} \mid \rvx_t] = \frac{1}{2} \frac{D_{\text{data}} \nu^2(t)}{\sigma^2(t)}~.
\end{equation}
{In} this case, the~score function suppresses entropy reduction in the subspace orthogonal to the data and therefore acts as a linear analog filter. Note that the entropy rate is zero when $D_\text{data}$ is equal to zero since all the distribution is in this case collapsed into a single point and no 'decision' needs to be~made. 


\section{Statistical Physics, Order Parameters and Phase~Transitions}
In this section, we will connect the information-theoretical concepts we outlined above with concepts from statistical physics such as order parameters, phase transitions and spontaneous symmetry breaking. We will start by studying the paths of fixed-points of the score function and use them to track 'generative decisions' (i.e., bifurcations) along the denoising trajectories. As~we will see, the~stability of these fixed-points paths is regulated by the Jacobian of the score and it is deeply connected with the conditional entropy production.

\subsection*{Branching 
 Paths of Fixed-Points and Spontaneous Symmetry~Breaking}
The fixed-points of the score function are defined by the equation
\begin{equation}
    \nabla \log{p_t(\x{t}^*)} = \vect{0}.
\end{equation}
We denote the set of fixed-points at time $t$ as $\Psi_t$. The~solutions of this fixed-point equation can be organized in a set $\Omega$ of piecewise continuous paths $\vect{\gamma}: \mathbb{R}^{+} \rightarrow \mathbb{R}^d \in \Omega$. To~remove ambiguities, we assume that, if~$\vect{\gamma}(\tau)$ is discontinuous at $\tau_0$, then the one-sided limit exists and $\gamma(\tau_0)$ is equal to $\lim_{t \rightarrow \tau_0^{+}} \vect{\gamma}(t) = \argmin_{\vect{x} \in \Psi_{\tau_0}} \lim_{t \rightarrow \tau_0^{+}} \left[\norm{\vect{x} - \vect{\gamma}(t)} \right]$. We know that $\lim_{t \rightarrow \infty} \vect{\gamma}(t) = \vect{0}$ for all paths since the zero vector is the only fixed-point of the score of the asymptotic Gaussian distribution. Any two paths $\vect{\gamma}_1(t)$ and $\vect{\gamma}_2(t)$ can be proven to overlap for a finite range of time, meaning that $\vect{\gamma}_1(t) = \vect{\gamma}_2(t)$ if $t \geq \tau_{1,2} \in \mathbb{R}^{+}$ (this follows from the results in \citep{carreira2002mode, carreira2003number, amendola2020maximum} on the number of modes of mixture of normal distributions). We refer to $\tau_{1,2}$ as the branching time of the two paths. The~branching time of two paths of fixed points can roughly be interpreted as a decision time in the generative process, where the sample will be 'pushed' by the noise in either one or the other path during the reverse dynamics. It is therefore insightful to study the behavior of the paths at the branching times. In~general, this can happen if there is a discontinuous jump in a path $\gamma(t)$. Perhaps more interestingly, two paths can also branch continuously at a finite time. This can be studied by analyzing the Jacobian matrix of the score function:
\begin{equation}
    J_t(\x{t}^*) = \nabla^T \nabla \log{p_t(\x{t}^*)}.
\end{equation}
{We} call a path point $\vect{\gamma}(t)$ stable at time $t$ if $J_t(\vect{\gamma}(t))$ is negative-definite. We say that the path is stable if this is true for all $t \in \mathbb{R}^{+}$ except for a countable set of time points $t_j$ where the Jacobian is negative semi-definite. Now consider two stable paths $\vect{\gamma}_1(t)$ and $\vect{\gamma}_2(t)$ that branch continuously at time $\tau_{1,2}$. Given the asymptotic separation vector 
$$
\vect{v}_{1,2} = \lim_{t \rightarrow \tau_{1,2}^{-}} \frac{(\vect{\gamma}_2(t) - \vect{\gamma}_1(t))}{\norm{\vect{\gamma}_2(t) - \vect{\gamma}_1(t)}}~,
$$
it can be shown that $\vect{v}^T J_t(\vect{\gamma}(t)) \vect{v} < 0$ in a finite interval $(\tau_{1,2}, \tau_{1,2} + \epsilon)$ and that
$$
\lim_{t \rightarrow \tau_{1,2}^{+}} \vect{v}^T J_t(\vect{\gamma}_1(t)) \vect{v} = 0~,
$$
which implies that the second directional derivative of $D^2_{\vect{v}} \log{p_t(\x{t})}$ along $\vect{v}$ vanishes at the branching~point. 

Consider now a generative diffusion with an initial distribution given as
\begin{equation}
    p_0(\y) = \frac{1}{K} \sum_{j=1}^K \delta(\y^{(j)} - \y)~,
\end{equation}
with $K$ distinct data-points $\y^{(j)} \in \mathbb{R}^d$. In~this case, there are exactly $K$ distinct stable fixed-point paths $\vect{\gamma}_j(t)$, with~$\vect{\gamma}_j(0) = \y^{(j)}$. Again, any two paths branch at a finite time $\tau_{j,k}$. For~a given $t$, we can partition the set of data-points in equivalence classes, where two data-points $\y^{(j)}$ and $\y^{(k)}$ share the same class if their associated path coincide at $t$. Importantly, each equivalence class corresponds to an individual fixed-point, which allows us to associate each fixed-point $\vect{x}^* \in \Psi_t$ with a sub-set of data-points that are, using colorful language, fused together. More precisely, we can express the fixed-points as weighted averages of data-points obtained by solving the self-consistency equation:
\begin{equation}
    \vect{x}^* = \sum_{j=1}^K w_j(\vect{x}^*) \y^{(j)}
\end{equation}
where 
\begin{equation} \label{eq: weights}
w_j(\vect{x}) = \frac{e^{\left(- \norm{\y^{(j)}}^2/2 + \vect{x}^T \y^{(j)} \right) /\sigma^2(t)}}
{\sum_{k=1}^K e^{\left(- \norm{\y^{(k)}}^2/2 + \vect{x}^T \y^{(k)} \right) /\sigma^2(t)}}~.
\end{equation}
{Note} that this average has non-zero weight on all data points, which is why we cannot find the location of the fixed-point solely based on its equivalence class. However, usually the weights corresponding to data-points in the equivalence class will be substantially larger than the other weights and will therefore dominate the average. In~summary, we can interpret the set of fixed-points as a decision tree where each branching point roughly coincides with a split between two sets of data~points.

\begin{figure}[H]
    \centering
    \includegraphics[width=0.95\linewidth]{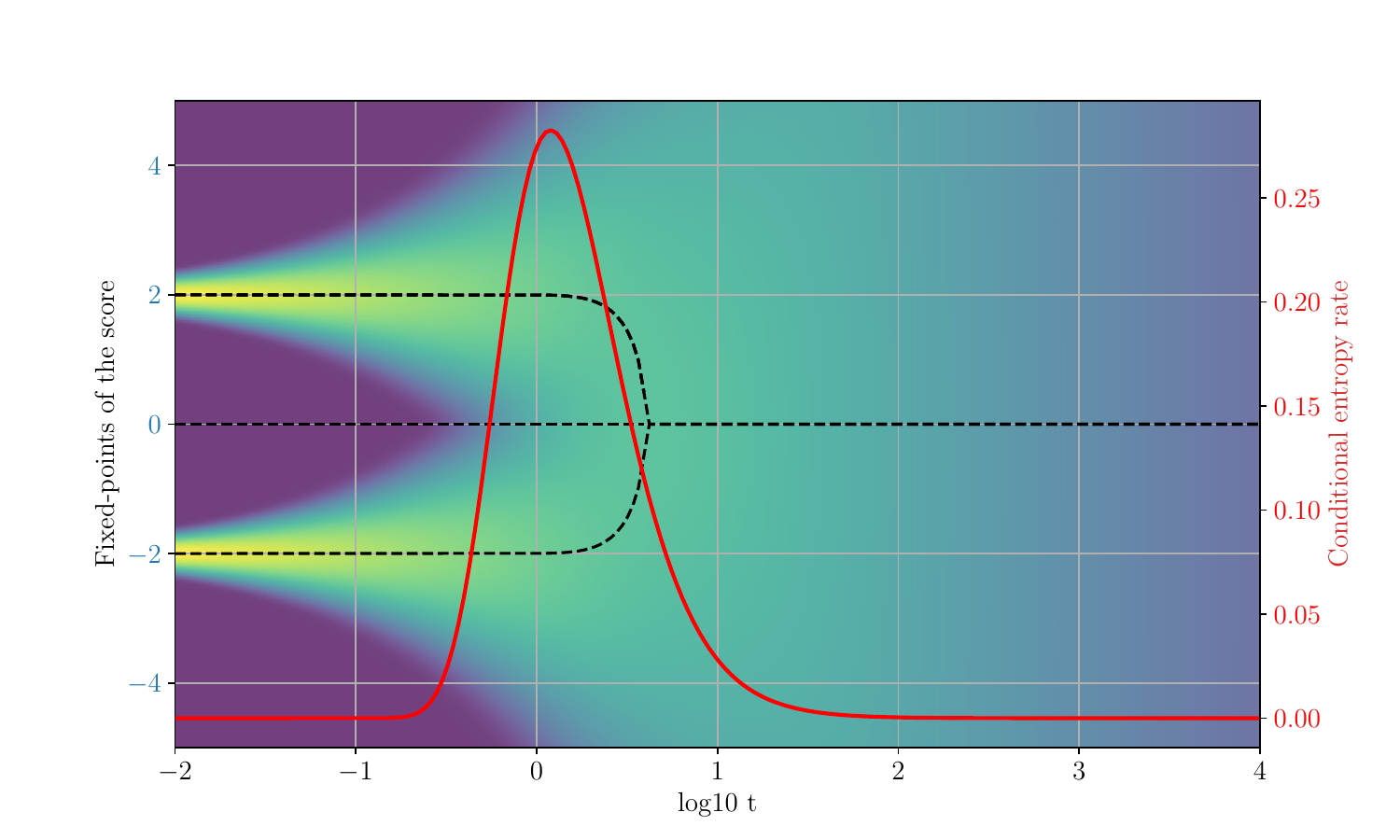}
    \caption{{Left}: Fixed points of the score field. Right: Conditional entropy rate. The~black dashed line denotes the stable fixed-point trajectories, while the red solid line represents the conditional entropy production rate. The~background color indicates the logarithm of the process~density.}
    \label{fig:entropy rate}
\end{figure}

\begin{figure}[H]
    \centering
    \includegraphics[width=0.65\linewidth]{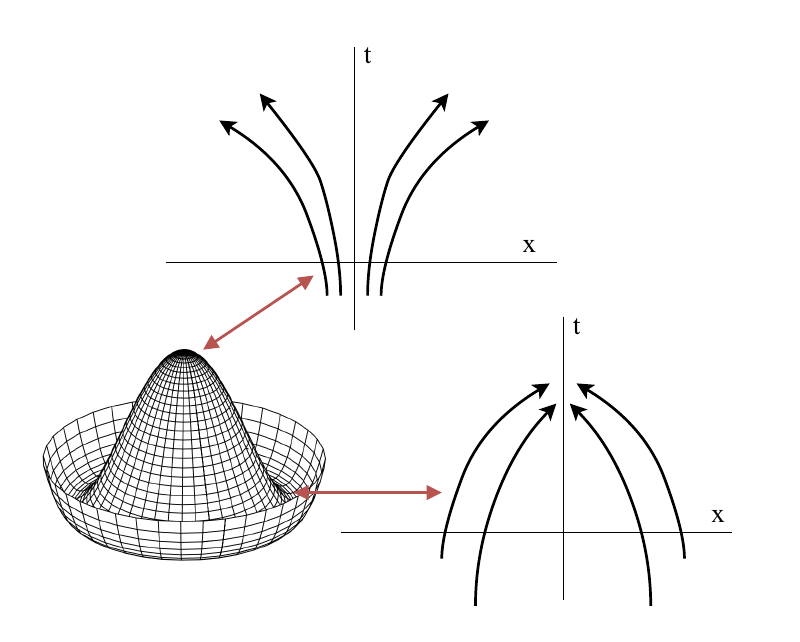}
    \caption{Stability 
 and instability of trajectories in different parts of a symmetry-breaking potential. Generative branching is associated with divergent~trajectories.}
    \label{fig:instability}
\end{figure}

An example of spontaneous symmetry breaking happens when the generative path needs to 'decide' between two isolated data-points. Consider again the mixture of the  delta case and two neighboring data-points $\y_1 = \vect{v}$ and $\y_2 = - \vect{v}$. If~the distance between the center of mass of these two points and the nearest external data-point is much larger than $\sigma(t)$, there will be a fixed point approximately located along the line segment connecting the two points. In~these conditions, we can consider the fixed-point equation restricted to the projections on $\vect{v}$:
\begin{equation} \label{eq: self-cponsistent equation}
    {x}_v^* = \tanh{\!\left(\frac{x_v^* + \phi(x_v^*, t) }{\sigma^2(t)} \right)}
\end{equation}
where $\phi(\vect{x}_v^*, t)$ encapsulates the interference due to all other data-points, which we, in~this example,   assume to be small relative to the norm of the separation vector:
$$
    \phi({x}_v^*, t) = \frac{\sigma^2(t)}{2} \left( \log{\left(e^{x_v^*} +  \sum_{j\neq 1,2}^K {y}^{(k)}_v e^{{\left(-\norm{\y^{(k)}}^2/2 + {x}_v y_v^{(k)} \right) /\sigma^2(t)}} \right)} - x^*_v \right)~.
$$
{If} we approximate the interference function with constant $\phi$ using a zero-th order Taylor expansion, Equation~(\ref{eq: self-cponsistent equation}) becomes the self-consistency equation of a Curie-=Weiss model of magnetism, with~temperature $T = \sigma^2(t)$ and external magnetic field $\phi$. The~solutions of this equation can be visualized as intersection points between a straight line and a hyperbolic tangent (see \citep{ambrogioni2024statistical,raya2023spontaneous} for a detailed analysis). When $\phi$ is finite, the~system transitions discontinuously from one to two fixed-points, which corresponds to a first-order phase transition in the magnetic system. However, the~size of the discontinuity vanishes when $\phi = 0$, when there is an exact symmetry between the two data-points (see Figure
~\ref{fig:entropy rate}). This gives rise to a so-called critical phase transition, where a single fixed point at $x^* = 0$ continuously splits into two paths $\vect{x}_1(t)$ and $x_2(t)$ with $\vect{x}_{1,2}(t - t_c) \sim \pm (t - t_c)^{1/2}$ for $t \rightarrow 0$. The~loss of stability of the fixed-point at the origin corresponds to the vanishing of the quadratic well around the point:
\begin{equation} \label{eq:stability condition}
    \frac{\partial^2}{\partial x_v^2} \log{p_{t_c}(x_{t_c}^*)} = 0,
\end{equation}
where, 
in~this case, $\x{t_c}^* = 0$ for $t < t_c$. The~analysis we just carried out involves the breaking of the permutation symmetry between two isolated data-points. On~the other hand, if~the symmetry is broken along all directions like in the case where the data manifold is a sphere centered at $\vect{x}_t^*$, Equation~(\ref{eq:stability condition}) implies that
\begin{equation} \label{eq: local trace vanishing}
    \text{Tr}\left[ \nabla^T \nabla \log{p_t(\vect{x}^*_{t_c})} \right] = \nabla \cdot \nabla \log{p_t(\vect{x}^*_{t_c})} = \vect{0}
\end{equation}
\textls[-15]{{Therefore,} the~change in stability condition can be reformulated as the local vanishing (or suppression in a less symmetric case) of the divergence of the vector field that drives the generative dynamics. The~transition from the super-critical ($t > t_c$) and the sub-critical ($t < t_c)$ phases then corresponds to a sign change in the divergence of the vector field \linebreak  (i.e., the~score) in the spherically symmetric case, or~a sign change of the divergence restricted to a sub-space in the general case, with~the sub-critical regime being characterized by positive eigenvalues of the Jacobi matrix that lead to divergent local trajectories (see Figure~\ref{fig:instability}). }
\unskip
\section{Dynamics of the Generative~Trajectories}
\label{Jacobian dynamics}

Around a point $\vect{x}_t$, the~local behavior of the generative trajectories under the deterministic ODE flow dynamics can be characterized by the eigenvalues of the symmetric part of the vector field, which quantify the separation rate of infinitesimally close~trajectories.

Let $\dot{\vect{x}} = f(\vect{x})$ and denote by $\x{t+\tau}(\cdot)$ the flow map that advances an initial condition at time $t$ by an increment $\tau$. For~a small perturbation $\vect{w}$ at $\vect{x}_t$, define the path difference
\begin{equation}
    \delta\vect{x}_t(\tau) := \x{t+\tau}(\vect{x}_t + \vect{w}) - \x{t+\tau}(\vect{x}_t).
\end{equation}
In the limit of vanishing $\|\vect{w}\|$ and small $\tau$, a~first-order expansion yields
\begin{equation}
    \delta\vect{x}_t(\tau) \approx \left(I + J_t \tau\right)\delta\vect{x}_t(0), 
    \qquad 
    J_t := Df(\vect{x}_t),
\end{equation}
and therefore the perturbation obeys
\begin{equation}
    \left.\frac{d\,\delta\vect{x}_t}{d\tau}\right|_{\tau=0} = J_t\,\delta\vect{x}_t(0).
\end{equation}
The evolution of the perturbation magnitude is controlled by the symmetric part of $J_t$ (note that for the gradient flows, like in the case of the diffusion, the~Jacobian is fully symmetric). Letting $S_t := \tfrac12(J_t + J_t^\top)$, we obtain
\begin{equation}
    \left.\frac{d}{d\tau}\|\delta\vect{x}_t(\tau)\|^2\right|_{\tau=0} = 2\,\delta\vect{x}_t(0)^\top S_t\,\delta\vect{x}_t(0).
\end{equation}
Maximizing over directions gives the local maximal stretching rate $\lambda_{\max}(S_t)$, while individual directions $\vect{w}$ experience the direction-dependent stretching rate given by the equation above. When the flow is integrated backward in time (as in reverse-time generation), these instantaneous stretching rates flip sign, so directions that are locally contracting forward in time become locally expanding under the reverse dynamics, leading to exponential sensitivity to perturbations over sufficiently short~horizons.

To connect this local stretching picture to symmetry breaking at $t_c$, consider the reverse-time generative dynamics in the immediate sub-critical regime $t=t_c-\epsilon$ with $\epsilon\ll 1$ and~linearize around the symmetric (unstable) branch $\vect{x}^*_t$. Writing the perturbation dynamics as $\dot{\delta\vect{x}}=-J_t(\vect{x}^*_t)\,\delta\vect{x}$, we obtain the finite-horizon propagation
\begin{equation}
    \delta\vect{x}(\tau)=e^{-\tau J_t(\vect{x}^*_t)}\,\delta\vect{x}(0),
\end{equation}
so that perturbations with overlap on expanding modes of the reverse dynamics grow exponentially over $\tau$ (equivalently, along directions that are locally contracting in the forward-time flow, i.e.,~modes associated with negative eigenvalues for the forward dynamics). In~the immediate sub-critical phase of a symmetry-breaking phase transition, we know that there is a non-empty subspace spanned by the eigenvector of the Jacobian corresponding to negative eigenvalues. Therefore, perturbations along this unstable eigen-space will be exponentially amplified by the generative dynamics. In~the stochastic case, this can be seen as a critical 'macroscopic amplification' of the infinitesimal noise input, where the noise breaks the symmetry of the generative model. In~the deterministic dynamics, the~symmetry is instead broken by the amplification of small differences between the generative~trajectories. 

In general, we will refer to the spectrum of Jacobian eigenvalues $\lambda_{j}(\vect{x}_{t}, t) $ as the local Lyapunov spectrum. As~we shall see, this spectrum can be directly related to the conditional entropy~production.

\subsection{The Global Perspective on Generative~Bifurcations}
In the previous sections, we characterized the generative dynamics of diffusion models by studying the associated paths of fixed points in term of their stability and bifurcations, which led us to establish formal connections with the statistical physics of phase transitions and symmetry breaking. However, in~high dimensions, small volumes around a fixed point have vanishingly low probability of being visited. In~fact, due to the dispersive effect of the noise, the~generative trajectories are concentrated on fixed-variance shells around the fixed points. More formally, these set of "typical" points form tubular neighborhoods of the set of fixed-points (see Figure 
 \ref{fig:entropy rate}). It is therefore unclear how a bifurcation in a path of fixed-points affects the behavior of the generative trajectories, since the analysis we presented in the previous sections was purely~local. 

To gain insight into the global behavior of the typical generative trajectories, we can study the expected divergence of the vector field at time $t$
\begin{equation}
    \text{div}(t) = \mean{ \nabla \cdot \score{t}}{\x{t}} = \mean{\text{Tr}\left[ \nabla^T \score{t} \right]}{\x{t}} ~.
\end{equation}
If $\text{div}(t)$ is negative, the~separation between the generative trajectories will, on~average, be contracted by the generative dynamics. The~simplest example of this contractive behavior can be studied by considering a data distribution with a single point: $p_0(\x{t}) = \delta(\y - \vect{c})$. In~this case, all trajectories converge to $\vect{c}$ for $t \rightarrow 0$, and~we have
\begin{equation}
    \text{div}_1(t) = -\frac{D}{\sigma^2(t)}~.
\end{equation}
where $D$ is the dimensionality of the space. In~the reverse dynamics, the~negative sign implies that the forward process produces a stable dynamics where the particles 'fall' towards the data~points. 

In the general case, this quantity can be identified with the 'trivial component'  of the expected divergence since it does not depend on the data but only on the forward process. In~the general case, it can be expressed as
\begin{equation}
    \text{div}_1(t) = \mean{\text{Tr}\left[ \nabla^T \nabla \log{p_t{(\x{t}\mid \y)}} \right]}{\x{t}}~.
\end{equation}
We can therefore study the purely data-dependent part of the expected divergence by subtracting this 'trivial component':
\begin{equation}
    \Delta \text{div}(t) = \text{div}(t) - \text{div}_1(t)  ~.
\end{equation}
Intuitively, $\Delta \text{div}(\x{t})$ encodes the separation of the typical trajectories in the reverse process due to bifurcations in the generative process, which mirrors the local analysis we carried out in the previous sections at the level of the~fixed points.

Using integration by parts, it is straightforward to connect the expected divergence with the conditional entropy rate 
\begin{equation}
    \dot{\mathbf{H}}[\y \mid \rvx_t] = \frac{\nu^2(t)}{2}  \Delta \text{div}(t)~ 
\end{equation}
Therefore, the~expected data-dependent divergence of the generative trajectories directly determines the conditional entropy rate. From~this identity, we can immediately deduce that $\Delta \text{div}(\x{t})$ is non-negative-valued and consequently that $\text{div}(t) \geq \text{div}_1(t)$. 

We can also show that the marginal entropy is produced by the expected divergence
\begin{equation}
    \dot{\mathbf{H}}[\rvx_t] = - \frac{\nu^2(t)}{2} \text{div}(t)~,
\end{equation}
which implies that $\text{div}(t) \leq 0$ since the marginal entropy is a monotonically increasing function of $t$ under our forward process. This reflects the fact that the forward process always leads to a dispersion of the trajectories, regardless to the nature of the initial distribution. From~this, we can conclude that the maximum bandwidth is achieved when 
\begin{equation}
    \text{div}(t) = \mean{\text{Tr}\left[ \nabla^T \score{t} \right]}{\x{t}} \rightarrow 0~.
\end{equation}
This gives us a clear connection between the local vanishing of the Jacobian in spontaneous symmetry breaking (Equation~(\ref{eq: local trace vanishing})) with the expected vanishing that corresponds to saturation of the generative~bandwidth. 

\subsection{Information~Geometry}
The derivation in the previous sub-section suggests a deep connection between the information production and the geometry of the data manifold. We can further analyze this connection by using concepts from information geometry~\cite{amari2016information}. The~key connection is that the conditional entropy rate is in fact just the expected value of the trace of the Fisher information matrix, which can be defined as follows:
\begin{equation}
    \mathcal{I}_t(\x{t}) = -\mean{\nabla \nabla^T \log{p(\y \mid \x{t})}}{\y \mid \x{t}}~.
\end{equation}
This quantity quantifies the sensitivity of the posterior distribution $p(\y \mid \x{t})$ to changes in $\x{t}$ and can be interpreted as a natural metric tensor on the variable $\x{t}$. Using Bayes' theorem and our simplified forward process, the~expression can be rewritten as
\begin{equation}
    \mathcal{I}_t(\x{t}) = \sigma^{-2}(t) \left(I + \sigma^2(t) {J(\x{t})} \right)~,
\end{equation}
Geometric information such as the manifold dimensionality is encoded in the spectrum of this matrix~\cite{kadkhodaie2023generalization, stanczuk2024diffusion, ventura2024manifolds, achilli2025memorization}. 
The Fisher information metric provides information on the (local) manifold structure of the data $\y$ as seen through the lenses of the noisy state $\x{t}$. This is easy to see in the case where the data is Gaussian with covariance matrix $\Sigma_0$, which gives the~formula
\begin{equation}
    \mathcal{I}_t = \sigma^{-2}(t) I - \left(\Sigma_0 + \sigma^2(t) I \right)^{-1}~.
\end{equation}
When $\y$ is supported on a $D_{\text{data}}$ manifold, the~(degenerate) eigenvalue $\lambda_{||}$ corresponding to the orthogonal complement is equal to zero. On~the other hand, in~the flat limit, the~tangent eigenvalues become equal to $\Sigma_0^{-1}$. This implies that the dimensionality of the manifold is given by the dimensionality of the eigenspace corresponding to the eigenvalue \mbox{$\lambda_{\parallel} = \sigma^{-2}(t)$}. In~the general case, the~eigen-decomposition of $\mathcal{I}(\x{t})$ characterizes the local tangent structure of the manifold~\cite{stanczuk2024diffusion, ventura2024manifolds}.

We can now use these expressions to cast light on the geometry of entropy production. The~conditional entropy rate is directly related to the trace of the Fisher information matrix:
\begin{equation}
    \dot{\mathbf{H}}[\y \mid \x{t}] = \frac{1}{2} \nu^2(t) \mean{\text{Tr}[\mathcal{I}(\rvx_t)]}{\rvx_t}~,
\end{equation}
which reduces to Equation~(\ref{eq: manifold entropy rate}) in the linear manifold case we just considered. From~this perspective, it is clear that the reduction in bandwidth is the result of the suppression of the eigenvalues of $\mathcal{I}(\x{t})$. This can also be seen in the general case by re-expressing the entropy rate in terms of the expected eigenvalues of the Jacobi matrix:
\begin{equation}
    \dot{\mathbf{H}}[\y \mid \x{t}] = \frac{\nu^2(t)}{2 \sigma^2(t)} \left(D + \sigma^2(t) \sum_{j} \mean{\lambda_j(\rvx_t)}{}\right)~.
\end{equation}
This equation shows that the entropy production is directly regulated by the spectrum of expected local Lyapunov exponents, as~studied in our local~analysis. 

We can better understand this formula by rewriting it as follows:
\begin{equation}
    \dot{\mathbf{H}}[\y \mid \x{t}] = \frac{\nu^2(t)}{2} \sum_{j} \left(1/\sigma^2(t) + \mean{\lambda_j(\rvx_t)}{}\right)~.
\end{equation}
From this, we can see that conditional entropy production in an eigenspace is fully suppressed when $\mean{\lambda_j(\rvx_t)}{} = - 1/\sigma^2(t)$, which is the eigenvalue of the Jacobian of the conditional score under the isotropic forward~process.

\section{A Stochastic Thermodynamic~Perspective}
\label{Variance peak}

A central question in generative diffusion is how uncertainty about the clean sample $x_0$ is resolved as the model evolves from the noisy state $x_t$ toward the data manifold. As~argued throughout this paper, the~appropriate notion of inferential uncertainty is the previously discussed conditional entropy $\mathbf{H}[\rvx_0 \mid \rvx_t]$ and, more fundamentally, its pathwise realization. The~study of pathwise entropy is naturally motivated by ideas from stochastic thermodynamics. However, we believe that the commonly used entropy in stochastic thermodynamics \citep{seifert2005entropy, premkumar2024neural} is not the correct quantity for understanding generative dynamics. It measures the irreversibility of the forward diffusion, not the uncertainty relevant to generating a single~outcome.

\begin{figure}[H]
\centering
\begin{minipage}[t]{0.48\linewidth}
\centering
\includegraphics[width=\linewidth]{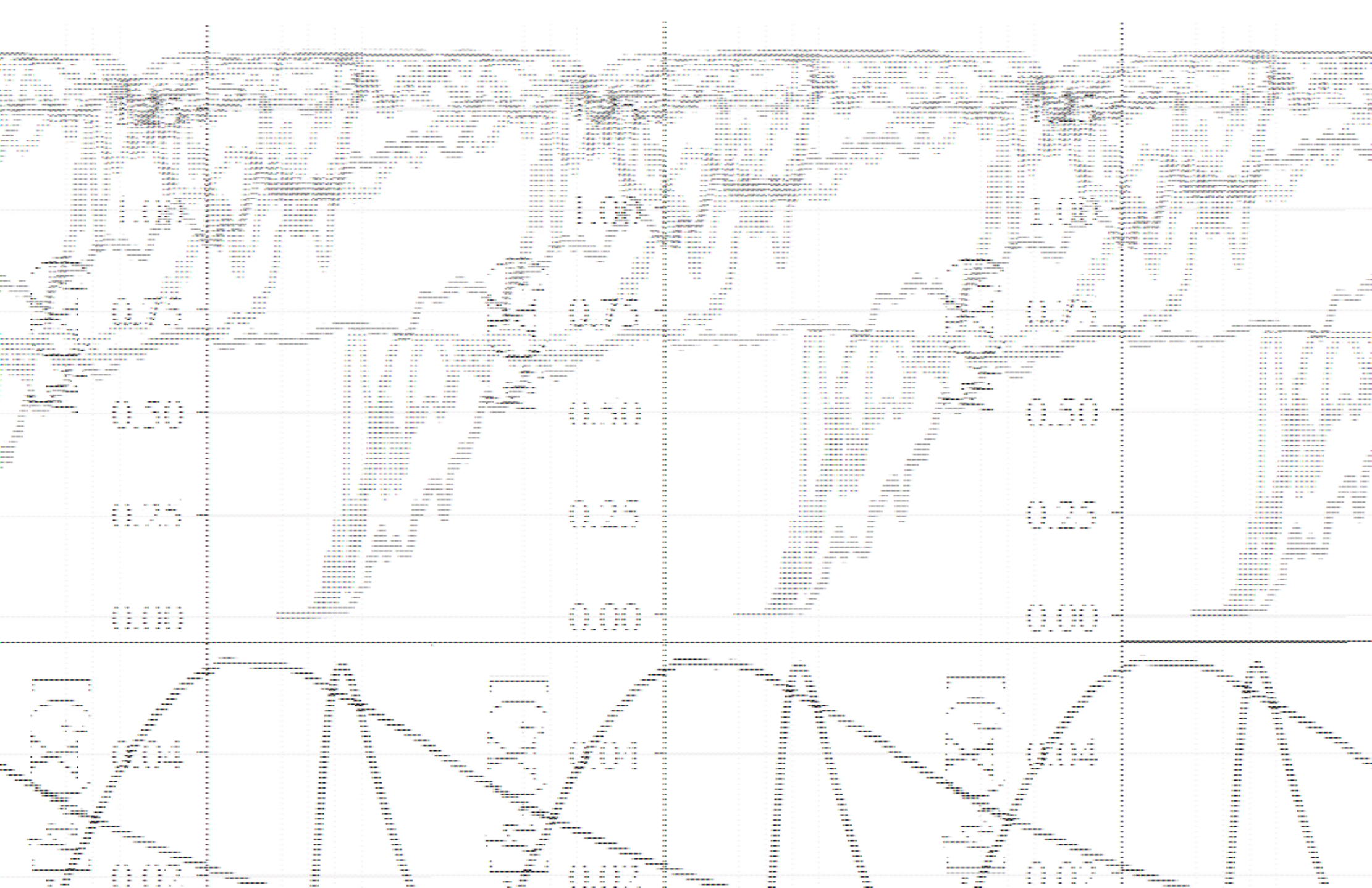}
\end{minipage}\hfill
\begin{minipage}[t]{0.50\linewidth}
\centering
\includegraphics[width=\linewidth]{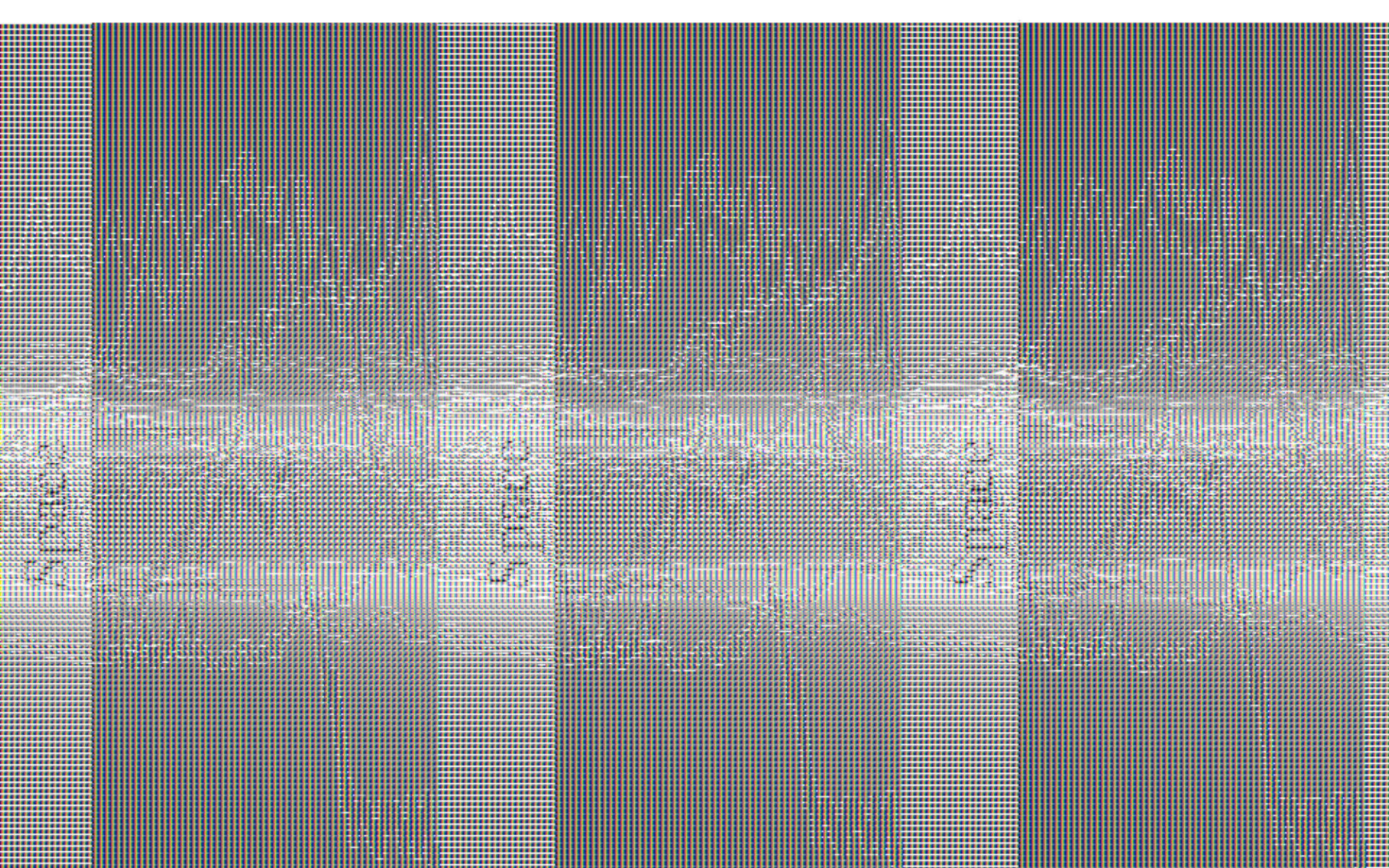}
\end{minipage}
\caption{An 
example of pathwise conditional entropies and corresponding paths for a four-point dataset located at $+2.3$, $+1.7$, $-1.7$, and $-2.3$. Each path is plotted in a distinct color, and the corresponding conditional-entropy curve uses the same color for visual matching. The background shading on the right panel depicts the normalized marginal density of the underlying SDE, with darker regions indicating lower probability.}
\label{fig: path-example}
\end{figure}

For a given point on the trajectory $\rvx_t$, we define its path-dependent conditional entropy~as
\begin{equation}
    h_t(\rvx_t)
    = -\int p(x_0 \mid \rvx_t)\,\log p(x_0 \mid \rvx_t)\,dx_0.
    \label{eq:path_entropy_def}
\end{equation}
This quantity measures the uncertainty experienced along a single generative path. Its expectation is the usual conditional entropy,
\[
\mathbb{E}[h_t(\rvx_t)] = H[\rvx_0 \mid \rvx_t],
\]
but its fluctuations encode a structure that is invisible to marginal entropies. In~particular, as~illustrated in Figure~\ref{fig: path-example}, the~pathwise conditional entropy $h_t(\rvx_t)$ can locally increase along individual generative trajectories even as the mean conditional entropy decreases, a~behavior reminiscent of entropy fluctuations in stochastic thermodynamics. Such effects do not arise in autoregressive models, where each generation step reduces uncertainty about the final sequence by revealing one token, since $\mathbf{H}[\rvx_{i+1:n}\mid\rvx_{1:i}] \leq \mathbf{H}[\rvx_{i+1:n}\mid\rvx_{1:i}] + \mathbf{H}[\rvx_i\mid\rvx_{1:i-1}] = \mathbf{H}[\rvx_{i:n}\mid\rvx_{1:i-1}]$. Whether these entropy fluctuations in diffusion-based generation have any practical advantage, however, remains an open~question.

To expose this dynamical heterogeneity, we consider the variance of the pathwise conditional entropy,
\begin{equation}
\mathcal{V}_h(t)\;:=\;\Var\!\big[h_t(\rvx_t)\big].
\end{equation}

\vspace{-5pt}
\unskip

\subsection{Variance of Pathwise Conditional Entropy as a Signature of Symmetry~Breaking}

We find that the variance captures symmetry-breaking transitions (see Figure~\ref{fig: path-example}). To gain a better understanding, we explore the general behavior of the variance and
demonstrate a connection with the speciation time\citep{biroli2024dynamical}.

Two limits are immediate. At~very early times, the~noise scale is negligible compared to the curvature of the data manifold. The~posterior $p(x_0\mid x_t)$ is effectively confined to the local tangent plane, behaving as an isotropic Gaussian whose shape is determined solely by the intrinsic dimension $D_{data}$ (we assume that the dimension is uniform across the manifold). Because~the entropy of this Gaussian depends on $k$ and $t$ but is insensitive to the specific location on the manifold,
\begin{equation}
h_t(x_t)\approx \text{const}
\qquad\Rightarrow\qquad
\mathcal{V}_h(t)\approx 0.
\end{equation}
At very late times, the~diffusion has effectively mixed the data distribution: $x_t$ carries little discriminative information about the origin $x_0$ and the posterior becomes approximately independent of $x_t$, again implying
\begin{equation}
h_t(x_t)\approx \text{const}
\qquad\Rightarrow\qquad
\mathcal{V}_h(t)\approx 0.
\end{equation}
Thus, nontrivial variance can only arise in an intermediate regime where different trajectories resolve uncertainty in different~ways.

Furthermore, near~a bifurcation/decision time $t_c$, the~ensemble contains a substantial fraction of trajectories that are already decisively committed to a branch and a substantial fraction that remain ambiguous. In~this regime, $h_t(x_t)$ becomes broadly distributed (some paths yield low entropy, others high entropy), and~$\mathcal{V}_h(t)$ is therefore~maximized.

\subsubsection*{Connection 
 with the Speciation~Time}

As already hinted at, for~Gaussian mixture models, the variance of the pathwise conditional entropy develops a pronounced peak on timescales of the order of the speciation time in the sense of \citet{biroli2024dynamical}. Interestingly, this behavior is observed for the VP SDE, but~is absent for the EDM and VE SDEs. We provide a brief, intuitive argument of the proof for the VP case here. More information about the setup and detailed derivation is deferred to Appendix \ref{appendix: proof}.

We consider a high-dimensional Gaussian mixture in the regime where, as~$d\to\infty$, the inter-class mean separations scale as $\|\mu_i-\mu_j\|=\Theta(\sqrt d)$ while within-class covariances remain $O(1)$. Throughout this discussion we focus on the Ornstein–Uhlenbeck (VP) forward process, since it yields a sharp speciation crossover. In~contrast, the~EDM parametrization does not generally exhibit the same sharp transition under an analogous~rescaling.

Introduce a latent class label $\rvz\in\{1,\dots,N\}$ so that $\rvx_0\mid(\rvz=i)$ is drawn from the $i$-th component. Conditioning on a realization $x_t$, the~conditional entropy satisfies
\begin{equation}
\mathbf{H}[\rvx_0\mid \rvx_t=x_t]
=
\mathbf{H}[\rvx_0\mid \rvx_t=x_t, \rvz]
\;+\;
\mathbf{H}[\rvz\mid \rvx_t=x_t]
\;-\;
\mathbf{H}[\rvz\mid \rvx_0, \rvx_t=x_t],
\label{eq:chain-rule-entropy}
\end{equation}
i.e., the chain rule for $(\rvx_0,\rvz)$ given $\rvx_t=x_t$. In~the well-separated regime (as is the case for $d\to\infty$), $\rvz$ is essentially fixed once $\rvx_0$ is known, so $H[\rvz\mid \rvx_0,\rvx_t=x_t]$ is negligible. For~Gaussian components and Gaussian forward kernels, the~first term is non-random, $p(x_0\mid x_t, \rvz=i)$ is Gaussian with a covariance that does not depend on $x_t$, and~in the shared-covariance case, it is the same across $i$. Thus the dominant source of fluctuations in $h_t(\rvx_t)$ is the class-uncertainty term $H[\rvz\mid \rvx_t]$. In~regions where the posterior concentrates on one component, $H[\rvz\mid x_t]\approx 0$, whereas in the maximally mixed region $H[\rvz\mid x_t]\approx H(\pi)$.

Under OU noising,
\begin{equation}
    \rvx_t=\alpha_t \rvx_0+\sigma_t\varepsilon,
    \qquad \alpha_t=e^{-t},\qquad \sigma_t^2=1-e^{-2t}.
\end{equation}
With $\|\mu_i-\mu_j\|=\Theta(\sqrt d)$ and $O(1)$ within-class covariance, the~posterior over $\rvz$ depends on $(t,d)$ essentially
through a single effective SNR,
\begin{equation}
\eta_d(t)\ :=\ \frac{d\,\alpha_t^2}{\sigma_t^2}
\;=\;\frac{d}{e^{2t}-1}.
\label{eq:eta-maintext}
\end{equation}
Equivalently, the~variance is a deterministic function that only depends on the effective~SNR,
\begin{equation}
\mathcal{V}_h(t)=\Var(h_t(\rvx_t))\ \approx\ F(\eta(t)),
\end{equation}
where $F$ is determined by the geometry of the mixture. The~endpoints are deterministic; $\eta\to\infty$ (early times) makes $\rvz$ effectively known and $h_t(\rvx_t)$ nearly constant, while $\eta\to 0$ (late times) makes $\rvz$ effectively independent of $\rvx_t$ and $h_t(\rvx_t)$ again nearly constant. Hence $F(\infty)=F(0)=0$, and~nontrivial variance can only occur when $\eta_d(t)=O(1)$.

\begin{figure}[H]
    \centering
    \includegraphics[width=0.55\linewidth]{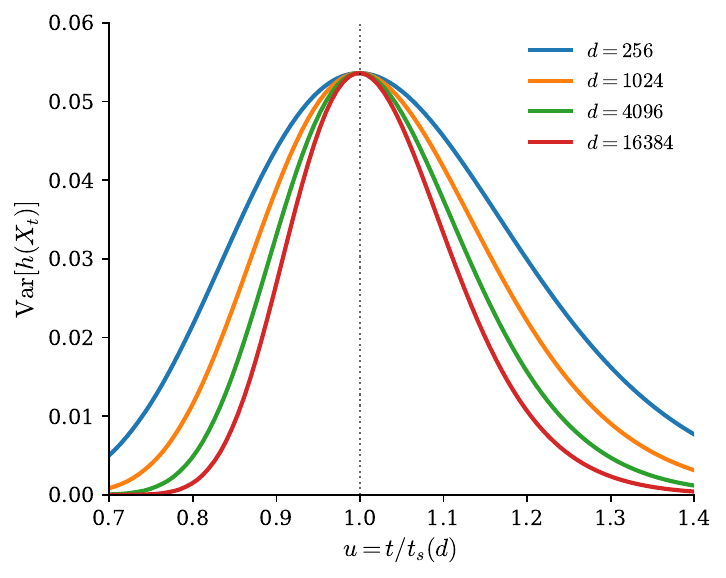}
    \caption{Evolution 
 of the variance of the pathwise conditional entropy as a function of time for a one-dimensional mixture of two Gaussians with means $\pm \sqrt{\frac{4}{3} d}$ and variances $1$.}
    \label{fig: asymptotics of var}
\end{figure}

This is exactly the speciation crossover. Using the dynamical-regimes rescaling \mbox{$t_S(d)=\tfrac12\log d$} and $u=t/t_S(d)$, we have
$e^{2t}=d^u$ and therefore
\begin{equation}
\eta_d(u t_S(d))=\frac{d}{d^u-1},
\end{equation}
which collapses to $\infty$ for $u<1$ and to $0$ for $u>1$, while staying $O(1)$ at $u=1$. Thus, as~$d$ grows, the~variance profile concentrates into a peak at $u=1$, matching the phase-transition behavior observed in Figure~\ref{fig: asymptotics of var}.

The same mechanism is not specific to Gaussians. What matters is (i) a latent `class' structure with separation $\Theta(\sqrt d)$ and (ii) $O(1)$ within-class fluctuations so that, after~the normalization implicit in \eqref{eq:eta-maintext}, component-specific details are subleading and the posterior over $\rvz$ is controlled by a single effective SNR. For~strongly non-Gaussian components, $\Var[h_t(\rvx_t)]$ may additionally show an early-time feature associated with rapid local Gaussianization under the forward kernel. This effect is absent in the ideal Gaussian case and is suppressed when each component is already close to~Gaussian.

\section{Discussion and Conclusions}

This paper has presented a unified framework that connects the dynamics, information theory, and~statistical physics of generative diffusion. We have shown that the generative process is governed by the conditional entropy rate, which is directly tied to the expected divergence of the score function’s vector field and, equivalently, to~the expected squared norm of the score. This quantity captures how uncertainty about the clean sample is resolved during denoising and reveals when the score is suppressed, allowing noise to drive the dynamics. In~this view, the~branching of generative trajectories arises from noise-induced symmetry-breaking transitions that occur when multiple datapoints remain compatible with the noisy state, and~the model is forced to commit to a specific~outcome.

By analyzing the fixed points of the score function and their stability, we showed that these generative decisions are formalized as bifurcations of the score field, which can be mapped onto classical symmetry-breaking phase transitions such as those described by mean-field models like the Curie--Weiss magnet. Peaks in the conditional entropy rate coincide with these bifurcation points, marking moments of maximal posterior mixing and heightened sensitivity to noise, where small fluctuations determine the generative branch taken by the~system.

Our results also clarify the relationship between generative diffusion and stochastic thermodynamics. While stochastic thermodynamic entropy characterizes the irreversibility of the process, the~conditional entropy studied here captures the inferential uncertainty relevant to generating a single sample. At~the trajectory level, the~pathwise conditional entropy and its variance reveal heterogeneity in how different generative paths resolve uncertainty, with~variance peaks emerging precisely during symmetry-breaking events. From~this perspective, entropy fluctuations are not incidental but constitute an information-theoretic signature of generative~decisions.

In conclusion, generative diffusion can be understood as a dynamical system that progressively breaks symmetries in the energy landscape while regulating the flow of information through posterior mixing. The~score function acts as a dynamic filter that suppresses noise along resolved directions while leaving unresolved directions weakly constrained, thereby controlling the generative bandwidth. This perspective provides a coherent explanation of how diffusion models transform noise into structured data and connects the learning dynamics of modern generative models to fundamental principles of information theory and statistical~physics.

\textls[-15]{Beyond conceptual unification, this framework suggests practical implications for model design and analysis. Because~entropy production and posterior mixing are directly linked to the score norm, they offer principled signals for identifying critical periods of high information transfer, motivating adaptive training and sampling strategies that target generative decision points \citep{stancevic2025entropic}. More broadly, the~information-thermodynamic perspective developed here provides a natural language for studying memorization, mode formation, and~generalization, and~may guide the development of future generative models that explicitly leverage controlled symmetry breaking to represent hierarchical and semantic~structure.}


\vspace{6pt} 





\authorcontributions{Conceptualization, D.S. and L.A.; methodology, D.S. and L.A.; software, D.S. and L.A.; validation, D.S. and L.A.; formal analysis, D.S. and L.A.; investigation, D.S. and L.A.; data curation, D.S. and L.A.; writing---original draft preparation, D.S. and L.A.; writing---review and editing, D.S.; visualization, D.S. and L.A.; project administration, L.A.; funding acquisition, L.A. All authors have read and agreed to the published version of the~manuscript.}

\funding{This research received no external~funding.}

\dataavailability{Data are contained within the article.}

\conflictsofinterest{The authors declare no conflicts of~interest.} 




\appendixtitles{yes} 
\appendixstart
\appendix
\section[\appendixname~\thesection]{Proof That the Variance Peaks at the Speciation Time}
\label{appendix: proof}

In this appendix, we show that, in~the asymptotic limit, the~variance of the pathwise conditional entropy develops a sharp peak at the speciation time. We begin by recalling the speciation scaling from the dynamical-regimes framework of \citet{biroli2024dynamical}. We then establish the claim in a minimal setting: a mixture of point masses (delta components), where the mechanism becomes completely transparent. Finally, we explain how the same argument extends to Gaussian mixtures. We focus on the point-mixture case because it isolates the core intuition with the least technical overhead. Furthermore, we use the ideas from the proof to show why the variance peak does not localize for the EDM~SDE.

\subsection{The Speciation~Time}

We follow the forward noising convention of \citet{biroli2024dynamical}, based on the Ornstein--Uhlenbeck (OU) process. In~closed form, the~forward marginal at time $t$ can be written~as
\begin{equation}
    \rvx_t = \alpha_t \rvx_0 + \sigma_t \varepsilon,
    \qquad \varepsilon\sim \mathcal N(0,I_d),
    \qquad \alpha_t=e^{-t},\ \ \sigma_t=\sqrt{1-e^{-2t}}.
    \label{eq:ou-closed-form}
\end{equation}
In the `dynamical regimes' picture, speciation occurs when the noise washes out the dominant class-structure mode. Denoting by $\Lambda$ the top eigenvalue of the data covariance, one obtains the criterion $\Lambda e^{-2t_S}=\Theta(1)$, hence $t_S=\tfrac12\log \Lambda + O(1)$. When $\Lambda\propto d$ (as in simple high-dimensional mixture models), this yields
\begin{equation}
    t_S(d)=\tfrac12\log d + O(1),
    \label{eq:tS-logd}
\end{equation}
and the transition becomes sharp in $O(1)$ windows around $t_S$ (in the original time variable~$t$). We show that the variance behaves the same and is governed by the same $t_s(d)$.

\subsection[\appendixname~\thesubsection]{Mixture of Data Points}
\label{appendix: mixture of data points}

Fix $N\ge 2$ and priors $\pi_i>0$ with $\sum_{i=1}^N \pi_i=1$. Choose distinct prototypes $m_1,\dots,m_N\in\mathbb R^N$. We consider the delta mixture supported on $N$ points whose separation grows like $\sqrt d$:
\begin{equation}
    \rvz\in\{1,\dots,N\},\quad \mathbb P(\rvz=i)=\pi_i,
    \qquad
    \rvx_0=\mu_Z^{(d)},\quad \mu_i^{(d)}:=\sqrt d\,m_i\in\mathbb R^N.
    \label{eq:means}
\end{equation}
We then apply the OU forward map \eqref{eq:ou-closed-form} in this $N$-dimensional space:
\begin{equation}
    \rvx_t = \alpha_t \sqrt d\, m_Z + \sigma_t G,
    \qquad G\sim\mathcal N(0,I_N).
    \label{eq:Yt}
\end{equation}
This is equivalent to embedding the same construction in a larger ambient $\mathbb R^d$ and noting that the extra orthogonal
coordinates carry only isotropic noise and cancel out   the class posterior; keeping only the $N$ signal coordinates is the
minimal~representation.

As described in the main text, using Equation 
 (\ref{eq:chain-rule-entropy}), the~problem simplifies to studying the posterior entropy random variable
\begin{equation}
    H_t^{(d)} := H(\rvz\mid \rvx_t)
    \equiv -\sum_{i=1}^N p(i\mid \rvx_t)\log p(i\mid \rvx_t),
\end{equation}
and its variance $\mathcal{V}^{(d)}_Z(t):=\Var(H_t^{(d)})$.

For the delta mixture, Bayes' rule gives
\begin{equation}
    p(i\mid x)
    \ \propto\ \pi_i \exp\!\left(-\frac{1}{2\sigma_t^2}\|x-\alpha_t \sqrt d\, m_i\|^2\right).
    \label{eq:posterior-bayes}
\end{equation}
Expanding the quadratic and dropping the $i$-independent term $-\|x\|^2/(2\sigma_t^2)$ yields a softmax with logits
\begin{equation}
    \ell_i(x)
    = \frac{\alpha_t\sqrt d}{\sigma_t^2}\, m_i^\top x
      -\frac{\alpha_t^2 d}{2\sigma_t^2}\,\|m_i\|^2,
    \qquad
    p(i\mid x)=\frac{\pi_i e^{\ell_i(x)}}{\sum_{j=1}^N \pi_j e^{\ell_j(x)}}.
    \label{eq:logits-x}
\end{equation}
Now, evaluate these logits at the random forward sample $x=\rvx_t$ from \eqref{eq:Yt}.
The only $(t,d)$-dependent combination that survives is the effective SNR
\begin{equation}
    \eta_d(t) := \frac{d\alpha_t^2}{\sigma_t^2}
    = \frac{d e^{-2t}}{1-e^{-2t}}
    = \frac{d}{e^{2t}-1}.
    \label{eq:eta}
\end{equation}
Indeed, substituting $\rvx_t=\alpha_t\sqrt d\,m_Z+\sigma_t G$ into \eqref{eq:logits-x} gives the random logits
\begin{equation}
    \ell_i(\rvx_t)
    = m_i^\top\!\big(\eta_d(t)\, m_Z + \sqrt{\eta_d(t)}\,G\big)
      -\frac{\eta_d(t)}{2}\|m_i\|^2,
    \label{eq:logits-eta}
\end{equation}
up to the additive constant $\log\pi_i$ in the softmax.
Crucially, all dependence on $(t,d)$ enters only through the single scalar $\eta_d(t)$; the remaining randomness is only $(\rvz,G)$. Therefore, the~law of $H(\rvz\mid \rvx_t)$ (and hence its variance) can be written as
\begin{equation}
    \mathcal{V}^{(d)}_Z(t)=\Var(H(\rvz\mid \rvx_t)) = F(\eta_d(t)),
    \label{eq:V=Feta}
\end{equation}
for a deterministic function $F:[0,\infty]\to[0,\infty)$ that depends only on $\{(\pi_i,m_i)\}_{i=1}^N$.

Taking the dynamical-regimes scaling
\begin{equation}
    t_S(d) := \frac12\log d,
    \qquad u := \frac{t}{t_S(d)},
    \label{eq:u-def}
\end{equation}
then $e^{2t}=e^{u\log d}=d^u$, and~\eqref{eq:eta} becomes
\begin{equation}
    \eta_d(u t_S(d)) = \frac{d}{d^u-1}.
    \label{eq:eta-u}
\end{equation}
As $d\to\infty$, this has the sharp dichotomy
\begin{equation}
\eta_d(u t_S(d))\to
\begin{cases}
\infty, & u<1,\\
0, & u>1,
\end{cases}
\qquad\text{while}\qquad
\eta_d(t_S(d))=\frac{d}{d-1}\to 1.
\label{eq:eta-limits}
\end{equation}
Now look at the two extreme-SNR limits of the posterior:

\begin{itemize}
\item Infinite SNR 
 ($\eta\to\infty$; early times $u<1$):
In \eqref{eq:logits-eta}, the~deterministic term $\eta\,m_i^\top m_Z$ dominates the Gaussian fluctuation $\sqrt{\eta}\,m_i^\top G$.
As a result, the~posterior concentrates on the true class $\rvz$ (with probability tending to $1$),
so $H(\rvz\mid \rvx_t)\to 0$ in probability. Hence $F(\infty)=0$.

\item Zero SNR ($\eta\to 0$; late times $u>1$):
In \eqref{eq:logits-eta}, all logits vanish, and~the posterior tends to the prior:
$p(i\mid \rvx_t)\to \pi_i$. Thus $H(\rvz\mid \rvx_t)\to H(\pi)$ deterministically, so $F(0)=0$.
\end{itemize}

Combining these endpoint behaviors with \eqref{eq:V=Feta} and \eqref{eq:eta-limits} gives the following collapse:
for every fixed $u\neq 1$,
\begin{equation}
    \mathcal{V}^{(d)}_Z(u t_S(d)) = F(\eta_d(u t_S(d))) \longrightarrow 0
    \qquad \text{as } d\to\infty,
\end{equation}
whereas at $u=1$,
\begin{equation}
    \mathcal{V}^{(d)}_Z(u t_S(d)) = F\!\left(\frac{d}{d-1}\right)\longrightarrow F(1).
\end{equation}
\textls[-35]{In other words, the~only place where the variance retains nontrivial structure is where $\eta_d(t)$ remains $O(1)$, and~under the rescaling $u=t/t_S(d)$, this occurs precisely at $u=1$ as $d \to \infty$. Equivalently, in~the original time variable $t$, the~peak lives in the  $O(1)$ window around $t_S(d)$.}

Furthermore, if~$F$ attains a maximizer at some finite $\eta^\star\in(0,\infty)$, and~if  $u_d^\star$ solves $\eta_d(u_d^\star t_S(d))=\eta^\star$, then using \eqref{eq:eta-u},
\[
\frac{d}{d^{u_d^\star}-1}=\eta^\star
\ \implies \
u_d^\star=\frac{\log(1+d/\eta^\star)}{\log d}
\to 1.
\]
Thus, regardless of the detailed shape of $F$, the~location of the peak converges to the speciation~point.

\subsection{Mixture of~Gaussians}

It turns out that for a mixture of Gaussians with the same covariances
\begin{equation}
    \rvx_0\mid (\rvz=i)\ \sim\ \mathcal N(\sqrt d\,m_i,\ \tau^2 I_N),
\end{equation}
repeating the above analysis yields the same conclusion as \eqref{eq:V=Feta}, but~with a renormalized effective SNR
\begin{equation}
    \tilde\eta_d(t)
    := \frac{d\alpha_t^2}{\tilde{\sigma}_t^2}
    = \frac{d e^{-2t}}{(1-e^{-2t})+\tau^2 e^{-2t}}
    = \frac{d}{e^{2t}-1+\tau^2}.
    \label{eq:eta-gauss}
\end{equation}
Since $\tau^2=O(1)$, the~same conclusion holds as for the mixture of data~points.

In the most general mixture of Gaussians, analysis becomes more convoluted, but~the same picture persists. Concretely, assume
\begin{equation}
    \rvx_0 \mid (\rvz=i)\ \sim\ \mathcal N(\sqrt d\,m_i,\ \Sigma_i),
    \qquad \|\Sigma_i\|=O(1)\ \text{as } d\to\infty .
    \label{eq:gmm-general}
\end{equation}
Under the OU forward map \eqref{eq:ou-closed-form},
\begin{equation}
    \rvx_t\mid (\rvz=i)\ \sim\ \mathcal N\!\big(\alpha_t\sqrt d\,m_i,\ \alpha_t^2\Sigma_i+\sigma_t^2 I_N\big).
    \label{eq:xt-general}
\end{equation}
\textls[15]{To expose the relevant scaling, it is convenient to normalize by the mean amplitude and~consider}
\begin{equation}
    \widetilde \rvx_t \ :=\ \frac{1}{\alpha_t\sqrt d}\,\rvx_t,
    \label{eq:xt-tilde}
\end{equation}
which is invertible for each fixed $t$ and therefore leaves the posterior (hence $H(\rvz\mid \rvx_t)$) unchanged:
$H(\rvz\mid \rvx_t)=H(\rvz\mid \widetilde \rvx_t)$. From~\eqref{eq:xt-general} we obtain
\begin{equation}
    \widetilde \rvx_t\mid (\rvz=i)
    \ \sim\ \mathcal N\!\Big(m_i,\ \frac{1}{d}\Sigma_i+\frac{\sigma_t^2}{\alpha_t^2\,d}\,I_N\Big).
    \label{eq:xt-tilde-law}
\end{equation}
Using the same effective SNR parameter as before,
\begin{equation}
    \eta_d(t):=\frac{d\alpha_t^2}{\sigma_t^2},
\end{equation}
the noise contribution becomes $\sigma_t^2/(\alpha_t^2 d)=1/\eta_d(t)$. Thus, after~normalization, each class-conditional can be viewed as the mean $m_i$ blurred by two noise sources: a component-specific term with covariance $\Sigma_i/d$ and a universal OU term with covariance $I_N/\eta_d(t)$,~i.e.,
\[
\widetilde \rvx_t \mid (\rvz=i) \ \sim\ \mathcal N\!\Big(m_i,\ \frac{1}{d}\Sigma_i+\frac{1}{\eta_d(t)}I_N\Big).
\]
The OU time affects the posterior only through the universal variance level $1/\eta_d(t)$, while the differences between components enter only through the $\Sigma_i/d$ term, which shrinks to zero as $d\to\infty$. Equivalently, for~each fixed $t$, the entropy variance can be written as a function of the single parameter $\eta_d(t)$, with~the dependence on the covariances entering
only through the rescaled matrices $\Sigma_i/d$:
\begin{equation}
    \mathcal{V}^{(d)}_Z(t)=\Var\!\big(H(\rvz\mid \rvx_t)\big)
    = \widetilde F\!\left(\eta_d(t);\ \left\{\Sigma_i/d\right\}_{i=1}^N\right),
    \label{eq:V=Feta-general}
\end{equation}
for some deterministic functional $\widetilde F$ determined by $\{(\pi_i,m_i,\Sigma_i/d)\}_{i=1}^N$.

Moreover, \eqref{eq:xt-tilde-law} makes the data-point limit explicit: for any fixed $\eta\in(0,\infty)$, we have $\Sigma_i/d\to 0$, so $\widetilde \rvx_t\mid(\rvz=i)$ converges to $\mathcal N(m_i,\, I_N/\eta)$ (i.e., it becomes a mixture of data points). Therefore, under~the dynamical-regimes rescaling $u=t/t_S(d)$ with $t_S(d)=\tfrac12\log d$, we still have $\eta_d(u t_S(d))=\frac{d}{d^u-1}$ and the sharp dichotomy between $\eta_d\to\infty$ for $u<1$ and $\eta_d\to 0$ for $u>1$. Combining these two facts yields the same conclusion as in the mixture of data points: for every fixed $u\neq 1$, $\mathcal{V}^{(d)}_Z(u t_S(d))\to 0$ as $d\to\infty$, while a nontrivial peak survives only at $u=1$ (i.e., at the speciation time).

\subsection{Variance Peak in the EDM~Case}

\begin{figure}[H]
\begin{adjustwidth}{-\extralength}{-4cm}
\centering
\subfloat[\centering  
]{%
  \centering\includegraphics[width=7.0cm]{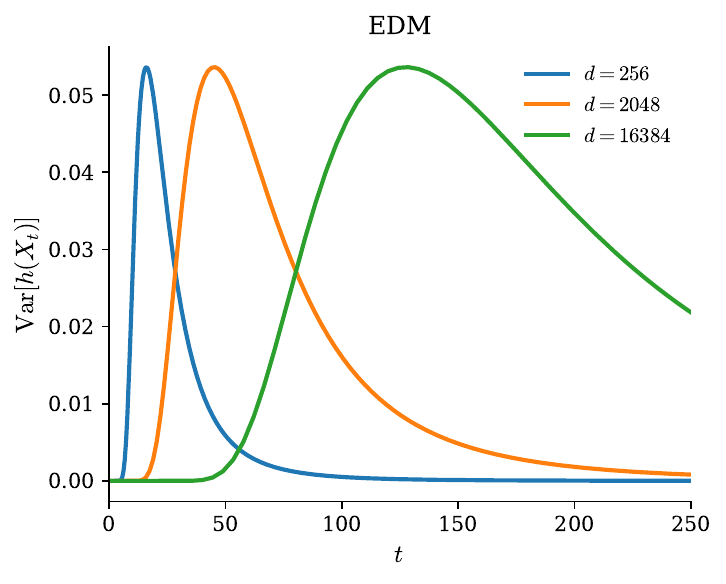}%
}
\subfloat[\centering ]{%
  \centering\includegraphics[width=7.0cm]{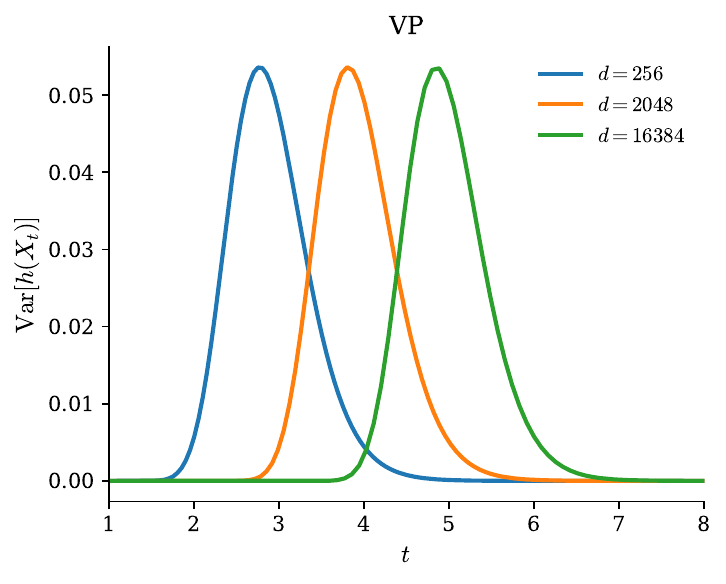}%
}
\end{adjustwidth}
\caption{{Variance} of the pathwise conditional entropy $h_t(\rvx_t)$ against time $t$ for equiprobable two-component Gaussian mixtures  for~(\textbf{a}) EDM and (\textbf{b}) VP SDEs and several values of $d$. In~EDM, the~transition region broadens with $d$, whereas in VP, it remains approximately constant in width, consistent with a transition that localizes in (rescaled) time for VP but not for~EDM.\label{fig: asymptotic_scaling_edm_vs_vp}}
\end{figure}

As stated in the main text, the~localization of the variance peak happens only for the VP SDE and not for the EDM/VP SDEs. To~better understand this, we can look at the proof in Appendix 
\ref{appendix: mixture of data points}. We see that the most of the proof does not depend on the forward process used. The~main mechanism through which the nature of the forward process plays a role is through an effective SNR, which in the EDM case is given by
\begin{equation}
    \eta_d(t) := \frac{d\alpha_t^2}{\sigma_t^2}
    = \frac{d}{t^2}.
    \label{eq:eta EDM}
\end{equation}
Now, by~requiring the effective SNR to be of order one (i.e., $\eta_d(t)=1$), we get
\begin{equation}
    t_S(d) := \sqrt{d},
    \qquad u := \frac{t}{t_S(d)}.
    \label{eq:u-def EDM}
\end{equation}
Therefore,
\begin{equation}
    \eta_d(u t_S(d)) = u^{-2}.
    \label{eq:eta-u EDM}
\end{equation}
Hence, the~effective SNR depends only on $u$ and not on $d$, which implies that the same holds for the variance
\begin{equation}
    \mathcal{V}^{(d)}_Z(u t_S(d)) = F(u^{-2}).
\end{equation}
This shows that there is no localization of the peak of the variance. Instead, as~$d$ increases, the~peak region broadens and scales with $\sqrt{d}$, rather than remaining of constant finite width as in the VP SDE. This difference between the two SDEs is shown in Figure~\ref{fig: asymptotic_scaling_edm_vs_vp}.

\vspace{-9pt}

\isPreprints{}{
\begin{adjustwidth}{-\extralength}{0cm}
} 

\reftitle{References}

\isPreprints{}{
} 
\end{document}